\documentclass[letterpaper, 10 pt, conference]{ieeeconf}  

\IEEEoverridecommandlockouts                              

\usepackage{graphics} 
\usepackage{graphicx}
\usepackage{flushend}

\title{\LARGE \bf
LiDAR-as-Camera for End-to-End Driving}

\author{Ardi Tampuu,
Romet Aidla,
Jan Are van Gent,
        Tambet Matiisen
\thanks{Manuscript accepted to Fresh Perspectives on the Future of Autonomous Driving workshop, hosted at ICRA 2022.}
\thanks{All authors are with the Institute of Computer Science,  University of Tartu, Estonia.}
\thanks{Code and instructions for getting access to data: https://github.com/UT-ADL/e2e-rally-estonia}
}

\begin{document}

\maketitle
\thispagestyle{empty}
\pagestyle{empty}

\begin{abstract}

The core task of any autonomous driving system is to transform sensory inputs into driving commands. In end-to-end driving, this is achieved via a neural network, with one or multiple cameras as the most commonly used input and low-level driving command, e.g. steering angle, as output. However, depth-sensing has been shown in simulation to make the end-to-end driving task easier. On a real car, combining depth and visual information can be challenging, due to the difficulty of obtaining good spatial and temporal alignment of the sensors. To alleviate alignment problems, Ouster LiDARs can output surround-view LiDAR-images with depth, intensity, and ambient radiation channels. These measurements originate from the same sensor, rendering them perfectly aligned in time and space. We demonstrate that such LiDAR-images are sufficient for the real-car road-following task and perform at least equally to camera-based models in the tested conditions, with the difference increasing when needing to generalize to new weather conditions. In the second direction of study, we reveal that the temporal smoothness of off-policy prediction sequences correlates equally well with actual on-policy driving ability as the commonly used mean absolute error.  

\end{abstract}

\section{INTRODUCTION}

\PARstart{F}{ully} end-to-end autonomous driving systems rely on a neural network to transform sensory inputs into raw driving commands, without clearly defined sub-modules \cite{ tampuu2020survey,ly2020learning,huang2020autonomous}. Most commonly, driving is achieved based solely on camera images \cite{bojarski2016end, osinski2019simulation, bansal2018chauffeurnet, kendall2019learning}, often with additional information about the desired route provided via one-hot encoded navigation commands or via route planner screen images \cite{codevilla2018end, sauer2018conditional, hawke2019urban}. This input combination is cheap and seems sufficient for human drivers to complete routes safely, making it an interesting object of study. However, in simulation a precise depth image can readily be generated and has been shown to be useful for driving models \cite{zhou2019does}. Even an approximate depth image predicted from RGB image may improve results \cite{zhou2019does,xiao2019multimodal}.   

In the real world, one can also attempt to predict depth images based on monocular camera images \cite{godard2019digging,alhashim2018high,yuan2022new}. Alternatively, stereo cameras can be used, but they suffer from a limited range. A more reliable depth image can be obtained by projecting the LiDAR point cloud to an RGB camera image. However, merging this depth image with a camera image is not trivial. The two sensors are usually not located in the same place and may see the world from different angles, leading to different blind spots. Besides the blind spots, extrinsic calibration of the sensors allows to match the depth and color values, but needs a precise procedure to be set up and may need to be repeated regularly. Furthermore, the capture frequencies may differ and it is not easy to guarantee temporal synchronization. Finally, even if good calibration is achieved for training data, synchronization failures or calibration errors may occur during deployment. 

To remove the need to use multiple sensors, Ouster LiDARs allow to generate a surround-view image containing perfectly aligned depth, intensity, and ambient radiation channels \cite{lidarascamera}. The intensity and ambient radiation channels can be seen as providing visual information, albeit from the infrared wavelength range. We, therefore, have a 3-channel input with temporally and spatially perfectly aligned visual and depth components.

This input is in image form and can be analyzed using any of the very successful approaches developed in computer vision. Network architectures for extracting information from images are more mature than network architectures for point clouds. We hence have a sensor providing rich information in a form that we know well how to analyze. 

This is the first experiment in our effort to validate the usefulness of these LiDAR-images for increasingly complicated driving tasks such as highway driving and urban driving. In here, we restrict ourselves to the simpler task of road following, albeit in the complex settings of real-world rally tracks chosen to be challenging also for humans. Our main contributions are the following:
\begin{enumerate}
    \item We compare the LiDAR-image-based driving with camera-based driving and show it yields beneficial robustness to light and weather conditions in this task.
    \item We study the correlation between off-policy and on-policy performance metrics, which has not been done before in the real car context.
    \item We collect and publish a real-world dataset of more than 500 km of driving on challenging rally tracks, with LiDAR and camera sensors and centimeter-level accurate GNSS trajectory. The dataset covers a diverse set of weather conditions, including snowy winter.
\end{enumerate}

\section{Methods}

\subsection{Behavioral Cloning}
Behavioral cloning takes a supervised learning approach to self-driving \cite{pomerleau1989alvinn}. Based on information from a chosen set of sensors, the model is optimized to produce the same driving behavior as a human would. This behavior is usually described by the sequence of low-level commands given or the trajectory taken \cite{tampuu2020survey,ly2020learning,huang2020autonomous}. For model training, a dataset is collected consisting of sensor recordings during human driving, accompanied by the driving commands or the trajectory produced by the driver. 

Such imitation approach has worked well for simpler tasks such as lane following \cite{bojarski2016end,pomerleau1989alvinn}, which seem to not require restrictive amounts of training data. However, dense traffic scenarios remain challenging for behavioral cloning \cite{codevilla2019exploring}. In addition to Tesla and comma.ai, multiple companies report promising performance in real-world urban driving with neural network based solutions \cite{hawke2019urban,jain2021autonomy, vitelli2021safetynet}, but it is unclear to what extent these can be considered as end-to-end. Though end-to-end models in other fields, e.g. speech recognition, have shown good generalization, replicating this success in self-driving is costly due to the massive amounts of data the cars produce. As further limitations, safety guarantees against rare situations and adversarial attacks are lacking \cite{kalra2016driving,nassi2020phantom} and interpreting model decisions remains challenging \cite{tampuu2020survey}. 

\subsection{Data Collection}
In the period of May 2021 to October 2021 training recordings of human driving were collected from all non-urban WRC Rally Estonia tracks and a few similar routes. Driving was performed with Lexus RX 450h fitted with a PACMod v3 drive-by-wire system provided by AutonomouStuff. The following sensors were recorded: NovAtel PwrPak7D-E2 GNSS device, Ouster OS1-128 LiDAR, three Sekonix SF3324 120-degree FOV cameras, and one Sekonix SF3325 60-degree FOV camera. All tracks were recorded in both directions at least once, amounting to more than 500 km of driving. The road type was mostly very low traffic gravel roads. There were shorter sections of two-lane paved roads. In January-February 2022 and in May 2022 further data collection was performed in snowy and early spring conditions. This data was only used for off-policy metric computation, not for training. The list of recordings used in this work are detailed in Appendix IV. 

The driving recordings from spring, summer, autumn, and winter differ strongly in vegetation levels and light conditions. All driving was done in daylight, but in differing weather conditions including heavy rain. The dataset, including recordings from sensors not used in this work, will be made fully available with this publication. 

\subsection{Data Preparation}
For neural network training, only recordings from Ouster OS1-128 mid-range LiDAR and Sekonix SF3324 RGB camera placed front center of the car were used. The list of recordings from May-October 2021 was divided into training (460 km of driving) and validation sets (80 km of driving). Recordings from the \emph{evaluation track}, where the on-policy evaluation was later performed, were not part of the training set, unless stated otherwise, but were part of the validation set that was used for early stopping. The lists of recordings used for model training and validations are given in Appendix IV.

\begin{figure}
    \centering
    \includegraphics[width=0.95\columnwidth]{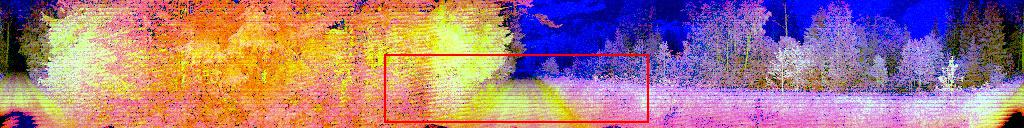}\\
    \vspace{0.1cm}
    \includegraphics[width=0.95\columnwidth]{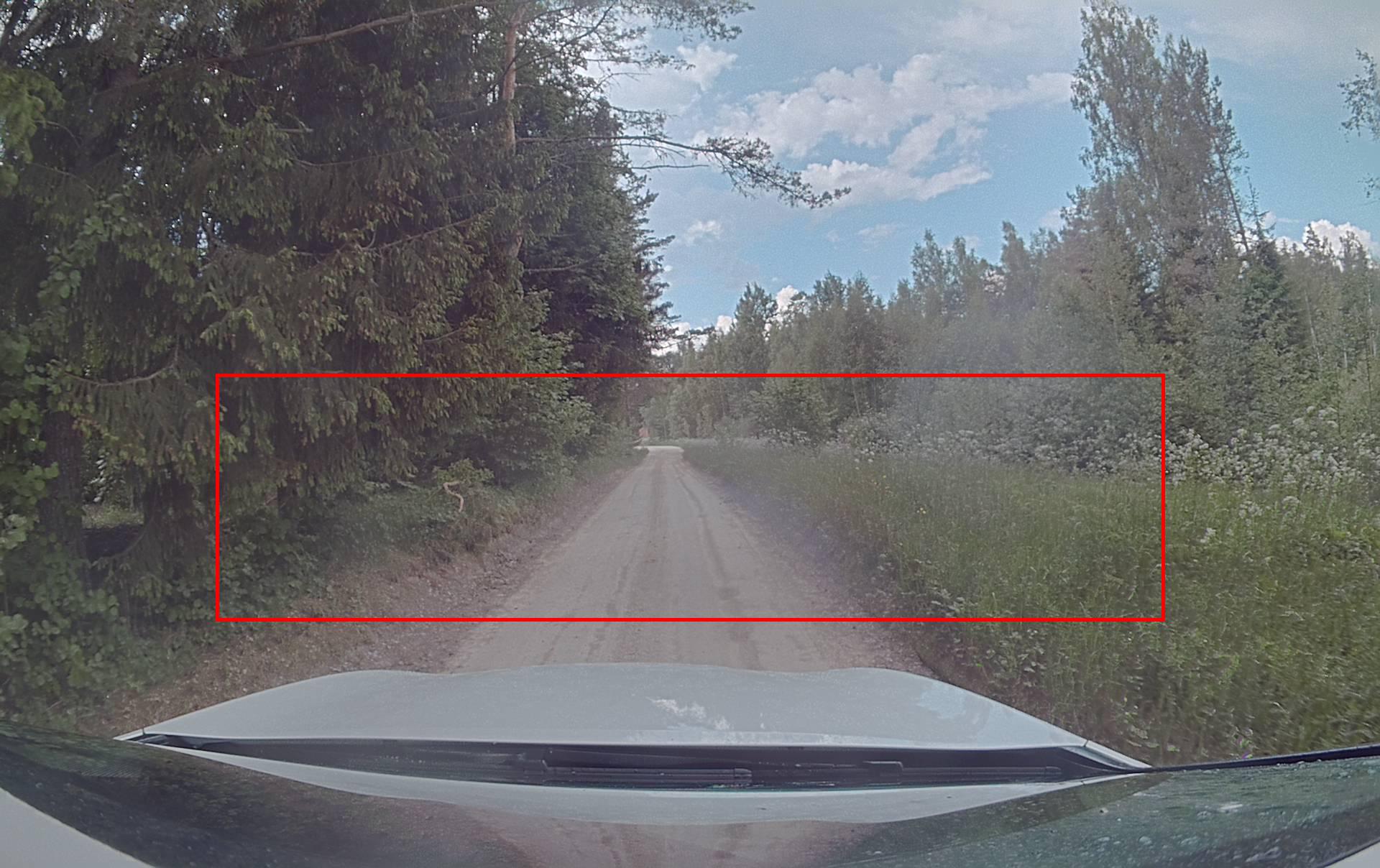}
    \caption{Input modalities. Red box marks the area used as model input. Top: surround view LiDAR image, with red: intensity, blue: depth and green: ambient. Bottom: 120 degree FOV camera.}
    \label{fig:inputs}
\end{figure}

The surround-view LiDAR-image output of Ouster OS1-128 LiDAR contains the channels of depth, intensity, and ambient radiation (Fig. \ref{fig:inputs}). In this work, the depth channel is further pre-processed in a way that distances in the range from 0 to 50 meters are mapped linearly to the values 255 to 0, i.e 20 cm depth resolution. All distances beyond 50 meters are marked as 0. For both RGB and LiDAR-image inputs, the image was cropped horizontally to remove the hood of the car and all rows above the horizon. For both input types, the image was cropped vertically to keep 90 degrees of view in the center front. The camera image was also downscaled to make it match LiDAR-image size. No further processing was done. This resulted in a 258x66x3 image as neural network input for both input types. The target labels correspond to the steering wheel angles as produced by human drivers.

No data diversification methods were employed. Firstly because we assume our data set is large enough to learn the task without augmentations. Secondly, useful augmentation is not easy to perform - recent works have described that models often learn to detect the fact of augmentation itself, instead of learning a generalized policy \cite{bojarski2020nvidia,commablog}.

The dataset was not balanced in any way. The rally tracks are curvy and and are not dominated by stereotypical drive-straight behaviour. We do not think there is any type of situation in our data that should be undersampled. Crossroads and interactions with other cars on the road are not excluded from the dataset, despite using the data only for learning road-following.

\subsection{Architecture and Training Details}
Here our intent is to use a relatively simple network architecture because the goal is to compare two input modalities rather than to achieve the best possible model. With limited data amount or variability powerful models could overfit, masking the effect of chosen inputs. Recent works have reported success on as low as 30 hours of training data \cite{hawke2019urban,codevilla2019exploring}. Our 500 km of data corresponds to only 15 hours of driving, placing us in danger of overfitting our models. 

We use a slightly modified version of the classical PilotNet architecture from \cite{bojarski2016end}. We add a batch normalization after each layer except the last two fully-connected layers. This gives us more training stability and faster convergence, similarly to what is discussed in \cite{santurkar2018does}. For similar reasons, we use LeakyReLU \cite{xu2015empirical} as the activation function instead of ReLU. The architecture is summarized in Table \ref{tab:architecture}. The model outputs the steering angle as the lateral control command. The driving speed is not controlled by the network.

\begin{table}[]
    \centering
    \caption{Architecture of the modified PilotNet network. Batch normalization always precedes activation function. All convolutions are applied with no padding.}
    \begin{tabular}{c|l|c|c}
Layer	& Hyperparameters	& BatchNorm &	Activation\\
\hline
Conv2d	& filters=24, size=5, stride=2&	BatchNorm2d&	LeakyReLU\\
Conv2d	&filters=24, size=5, stride=2	&BatchNorm2d&	LeakyReLU\\
Conv2d	&filters=24, size=5, stride=2	&BatchNorm2d&	LeakyReLU\\
Conv2d	&filters=48, size=5, stride=2	&BatchNorm2d&	LeakyReLU\\
Conv2d	& filters=64, size=3, stride=1&	BatchNorm2d&	LeakyReLU\\
Flatten	& -& -&	-	\\
Linear	&nodes=100	&BatchNorm1d&	LeakyReLU\\
Linear	&nodes=50	&BatchNorm1d&	LeakyReLU\\
Linear	&nodes=10	&none&	LeakyReLU\\
Linear	&nodes=1	&none&	none
    \end{tabular}
    \label{tab:architecture}
\end{table}

We use mean absolute error (MAE) as the loss function, as this metric has been shown to correlate better than mean squared error with on-policy driving ability \cite{codevilla2018offline}. We use Adam optimizer with weight decay \cite{loshchilov2017decoupled} with default parameters in PyTorch \cite{NEURIPS2019_9015}. We use early stopping if no improvement in the validation set was achieved in 10 consecutive epochs, with the maximum epoch count fixed to 100. The code with model definitions and training procedure will be made available in GitHub with the publication.

It has been reported that multiple training runs can result in clearly differing on-policy behaviors \cite{codevilla2019exploring}. To minimize the potential effect of training instability, we train three versions of our main models and report the metrics for each.

\subsection{Evaluation Metrics}
The models were evaluated on- and off-policy. It is widely reported that off-policy metrics correlate poorly with actual driving ability \cite{codevilla2018offline}. However, they are cheap to compute before deploying the solution. If a better off-policy metric could be found, development could be sped up by allowing to select only the best models for real-life evaluation. 

Off-policy metrics are computed using human-driven validation recordings originating from the same season as when the on-policy evaluation happened. We limit the set of recordings to the same season because we assume that off-policy metrics computed on summer data would have no information about driving ability in the winter and vice versa.

We report the mean absolute error (MAE) between human commands and model predictions as this metric has been reported as having favorable correlation with driving ability \cite{codevilla2018offline}. In addition, following \cite{eraqi2017end, fernandez2018two}, we also compute the whiteness of the predicted command sequence:\\
\begin{equation}
W=\sqrt{\frac{1}{D}\sum^D_{i=1}(\frac{\delta P_i}{\delta t})^2},
\end{equation}
where $\delta P_i$ is the change in predicted steering angle, D is the size of the dataset, and $\delta t$ the temporal difference between decisions. $\delta t$=0.1 for LiDAR and $\delta t$=0.033 for camera.

Whiteness measures the mean smoothness of the sequence of commands generated and can be computed on- and off-policy. In here, $W_\mathsf{{off-policy}}$ refers to the whiteness of the sequence of commands generated by a model on human-driving recordings from the evaluation track, in the same season as when the on-policy testing took place. In contrast, $W_\mathsf{{on-policy}}$ refers to the whiteness of the commands generated during model-controlled driving during evaluation.

We consider the smoothness of commands a promising metric because during on-policy testing we observed that jerkiness of driving, i.e. temporally uncorrelated commands, seems to predict an imminent intervention. Model not responding to consecutive very similar frames in a consistent manner might reveal its inability to deal with the situation. 

The number of interventions during a test route was counted and distance per intervention (DpI) computed as the main quality metric. The models were trained to perform route following and not to handle intersections. All interventions at intersections were removed from the count. For safety reasons, in the case of an oncoming car, the safety driver always took over the driving. Interventions due to traffic were also excluded from the intervention count.

As additional on-policy metrics, we measure the deviation of the model driving compared to a human trajectory on the same route. Locations were measured using NovAtel PwrPak7 GNSS receiver combining the inertial navigation system (INS) data with real-time kinematic positioning (RTK) achieving centimeter-level precision. For each position in the model-driven trajectory, the offset is defined as the average distance to the two closest human trajectory points. The mean of this lateral difference along the route is reported as $MAE_\mathsf{{trajectory}}$. We define $\mathsf{failure\ rate}$ as the proportion of time this lateral difference was above 1 meter.

\subsection{On-policy Evaluation Procedure}

On-policy evaluation was performed on a 4.3 km section of SS20/23 Elva track in both driving directions (cf. Appendix II for map of the route). The speed along the route was set to 80\% of the speed a human used in the same location on the route, as extracted from a prior recording of human driving. Driving at 100\% human speed was attempted, but was too dangerous to use with weaker models. For covering different weather conditions, it was intended to be completed in two parts: autumn and winter, but due to technical reasons, a third session in spring was needed.
\begin{itemize}
    \item In the last week of November 2021: the weather conditions and vegetation levels were very similar to the most recent training data recorded in the end of October. Due to a missing parameter in the inference code, the RGB models were run on BGR input and the results had to be discarded. Hence, only LiDAR-based models were adequately tested in this session. Night driving was performed with dipped-beam headlights on. Results from these tests are marked with (Nov) in the Table \ref{tab:main_results}.
    
    \item In the first week of February 2022: with snow coverage on the road. This constitutes a clearly out-of-distribution scenery for the camera models. Moreover, also for LiDAR models the surface shapes and reflectivity of snow piles differ from vegetation and constitute out-of-distribution conditions. LiDAR and camera images from summer, autumn and winter are given in Appendix I. From this trial, marked with (Jan), we report only driving performance with LiDAR, as camera still operated in BGR mode.
    
    \item In the first week of May 2022: early spring, which constitutes a close-to-training-distribution condition. Camera models were evaluated with correct inference code. The location of LiDAR on the car had changed before this trial compared to the training data. LiDAR-based models underperfomed during this test, despite our efforts to adjust the inputs.
\end{itemize}

In short, camera models were tested with adequate inputs only in spring 2022, whereas LiDAR models only in autumn 2021 and winter 2022. The weather conditions were not identical and direct comparison of values should not be made.

The rally tracks are narrow and bordered by objects harmful to the car, hence the safety driver was at liberty to take over whenever they perceived danger. An intervention is hence defined as a situation where the safety driver perceived excessive threat to the car or the passengers. An intervention was triggered by the safety driver applying force to turn the steering wheel. If the model turned the steering wheel at the same moment and in the same direction as the safety driver, no force was applied and no intervention counted.

\section{Results}
In this section, we present the driving ability of our models as measured by on-policy metrics. We also compute off-policy metrics, but only with the purpose to evaluate their correlations with on-policy performance.   

\subsection{Driving on an Unseen Track}
The ultimate goal of end-to-end self driving is to create models that can generalize to new roads without the use of high-definition maps. Hence, we first summarize the models' ability to generalize to the evaluation track from other similar roads. Three instances of LiDAR models were tested in autumn and three camera models in spring. Using multiple models allows the reader to grasp the stability of results. A larger number of repetitions was not used due to the complexity of real-world evaluation. The metrics for these six evaluations are given in the first section of Table \ref{tab:main_results}. 

\begin{table*}[]
\centering
\caption{Results of on-policy evaluations. Evaluations interrupted due to a high frequency of interventions are marked with *. Horizontal lines separate values illustrating the different Results subsections.}
\begin{tabular}{c|p{2.8cm}|c|c|c|c|c|c}
Experiment &Model (session) & Distance & Interventions & DpI & $MAE_\mathsf{{trajectory}}$ & Failure rate & $W_\mathsf{{on-policy}}$\\
\hline

Generalization&Camera v1 (May) &	8464.33m &	2 &	4232.17m &	0.2309m &	0.96\% &	24.63°/s\\
&Camera v2 (May) &	8363.17m &	4 &	2090.79m &	0.2382m &	0.69\% &33.46°/s\\
&Camera v3 (May) &8389.88m &7 &1198.55m &0.2403m &0.66\% &29.70°/s\\

&LiDAR v1 (Nov) & 8442.5m & 2 & 4221.3m & 0.22m & 0.42\% & 23.0°/s \\
&LiDAR v2 (Nov) & 8465.9m & 2 & 4233.0m & 0.24m & 0.98\% & 17.7°/s \\
&LiDAR v3 (Nov) & 8432.3m & 3 & 2810.8m & 0.25m & 2.18\% & 18.8°/s \\
\hline
Overfitting & Camera overfit (May) & 8489.14m &3 &2829.71m &0.2453m &1.39\% &	23.07°/s \\
&LiDAR overfit (Nov)& 8436.9m & 0 & $>$8436.9m & 0.26m & 4.38\% & 19.2°/s \\
\hline
Night&LiDAR v2 (Nov) & 8216.3m & 8 & 1027.0m & 0.24m & 1.27\% & 25.6°/s \\
&LiDAR v2 \#2 (Nov) & 8376.6m & 3 & 2792.2m & 0.23m & 1.55\% & 20.3°/s \\
&LiDAR overfit (Nov) & 8521.5m & 1 & 8521.5m & 0.25m & 1.52\% & 21.5°/s\\
\hline
Winter &LiDAR v1 (Jan) & 8080.5m & 19 & 425.3m & 0.24m & 0.94\% & 38.4°/s \\
&LiDAR v2 (Jan)  & 8001.4m & 22 & 363.7m & 0.28m & 3.10\% & 38.7°/s\\
&LiDAR v3 (Jan)  & 7698.9m & 34 & 226.4m & 0.26m & 1.64\% & 42.2°/s\\
\hline
LiDAR&LiDAR v2 (Nov)& 8491.6m & 0 & $>$8491.6m & 0.22m & 1.97\% & 19.2°/s\\
channels&LiDAR intensity (Nov)& 8446.2m & 2 & 4223.1m & 0.33m & 7.02\% & 24.0°/s \\
(next day) & LiDAR depth (Nov)& *1679.0m & *22 & *76.3m & *0.61m & *19.95\% & *29.9°/s \\
&LiDAR ambience (Nov)& *329.5m & *19 & *17.3m & *0.73m & *17.49\% & *168.2°/s \\
    \end{tabular}
    \label{tab:main_results}
\end{table*}

The results indicate that in in-distribution weather, but on a novel route, the performance of LiDAR-based models is similar or better than camera models. The evaluations took place half a year apart, but conditions were suitable in both cases. Spring testing was done in largely cloudy day, with only short periods of direct sunlight. Autumn test took place in cloudy and dim daylight with short periods of very light rain. These conditions should be sufficiently close to ideal for camera and LiDAR models respectively.


\subsection{Overfitting Setting}
We next asked to what extent the task was more difficult due to needing to generalize to a new route. We trained camera and LiDAR models that included, in their training set, one human driving recording in each direction from the evaluation track. As these models will be exposed to the objects and types of turns on the evaluation track, we call this the "overfitting" (to the evaluation track) setting. The second section of Table \ref{tab:main_results} indicate that while the LiDAR model clearly benefitted from test-track recordings, the effect is weaker for the camera-based model. Overfitted LiDAR model drove without interventions, while the RGB-model yielded similar performance to the non-overfit models. 

We conclude that approximately 500 km of road following data in the original training set did not suffice for good generalization to similar but unseen roads. Data augmentation techniques could be applied or more data collected to increase generalization over this source of data variability. 

\subsection{Night Driving and Winter Driving}
The third set of on-policy tests evaluates the models' ability to generalize to weather conditions very different from the training distribution. We have \emph{a priori} no expectation of camera-based driving models generalizing to these conditions. The camera images during the night differ drastically from daylight driving, despite using headlights. Similarly, the color distribution and brightness of camera images in the winter with snow coverage is clearly out-of-distribution. These differences are easy to detect for the human eye.

In contrast, the extent that these two novel conditions are out-of-distribution for LiDAR-based models is difficult to estimate by naked eye. \emph{A priori}, we can assume that depth and intensity channels should be affected only minimally by lack of sunlight in night driving, with ambient radiation somewhat affected. Snow coverage adds more smooth surfaces to the landscape, but the resulting depth image may remain in the proximity of the diversity of scenes contained in the training data. Ambient radiation and intensity images are likely out-of-distribution due different reflective properties of snow and vegetation, but the extent of its effect on LiDAR-based driving models was unknown before tested.

The results from these trials are marked with night and winter in Table \ref{tab:main_results}. The camera models immediately steered the car off the road, there is no performance to report. While the experiments were done with flawed BGR input to the models, judging from the performance of BGR models in other experiments we do not expect the performance with RGB input to be much different (cf. Appendix V). LiDAR models trained with day-time data sets see only a minimal drop in performance when deployed at night. 

However, when deploying models trained on data from spring, summer, and autumn to snow-covered roads, also LiDAR-based models see a clear drop in performance. LiDAR models manage to maintain some of their driving ability, but drop from $\approx$ 4000m to 226-425 meters per intervention. Qualitatively, we report that LiDAR models drove reasonably well in the forest where depth information was abundant, but failed to stay on the road on sections between open fields (cf. Appendix III).

\subsection{Informativeness of Individual LiDAR Channels}

We also performed on-policy testing of models trained on individual LiDAR-image channels. This was done to obtain a better understanding of the usefulness of each of these channels. This evaluation was performed in in-distribution weather, in November 2021. As these experiments were performed on another day compared to the tables above, a 3-channel model was also re-evaluated to confirm the conditions were similar. The tested models were trained with no recordings from the evaluation track in the training set.

The examples of the images from these three channels in summer, autumn, and winter are given in the Appendix II. At visual inspection, the intensity channel seems approximately as sharp and as informative as a gray-scale camera image, albeit capturing a different wavelength. The depth image is less spatially dense but clearly informative about sufficiently large obstacles. However, ambient radiation images depend strongly on sunlight being present and seem an unreliable source of information.  

In the Table \ref{tab:main_results} we observe that the model trained based on intensity channel can perform surprisingly well. However, neither depth nor ambient radiation channels contain sufficient information for safe driving. Depth-based model struggled to drive safely also in the forest, where trees could have provided depth-cues of where to steer towards. These channels may nevertheless still contribute useful information to the 3-channel model.

\subsection{Correlation Study Between On- and Off-Policy Metrics}


In this work, we trained a total of 11 models (3+1 LiDAR, 3+1 camera, and individual LiDAR channels). We deployed these models in more and less suitable conditions, including accidentally deploying RGB-image models using BGR video stream and deploying LiDAR models after the sensor location had been changed. Here, we wished to study if the on-policy performance of these test-drives could have been predicted before deployment via off-policy metrics or at least during the drive via non-discrete on-policy metrics.

In the following we used metrics from 17 model-deployments. The list of trials included and the associated metrics are given in Appendix IV. This list includes using RGB-based model with BGR camera stream and using LiDAR in changed location, because the models were capable of driving despite the disadvantageous conditions. Each deployment was matched with an off-policy evaluation using recordings from the same track, similar season and similar sensor configuration (e.g. using BGR images). We computed the Pearson correlation \cite{hauke2011comparison} between DpI and the various other on- and off-policy metrics. The DpI for trials with no interventions was set to 10 km for the computations. The resulting correlation values are given in Table \ref{tab:correlations}.

\begin{table*}
\caption{Pearson correlations of the main driving quality metric distance per intervention (DpI) with other on- and off-policy metrics. The highest-correlating metrics of both types are highlighted in bold.}
\centering
\begin{tabular}
{c|c|c|c|c||c|c}
&\multicolumn{4}{c||}{On-policy measures}&\multicolumn{2}{c}{Off-policy measures}\\
Measure & $MAE_\mathsf{{trajectory}}$ & Failure rate &$W_\mathsf{{on-policy}}$ & $W_\mathsf{{effective}}$ & $W_\mathsf{{off-policy}}$ & $MAE_\mathsf{{steer}}$ \\
\hline
Pearson R & -0.56 & -0.06 & -0.56 &  \textbf{-0.67} & -0.72 & \textbf{-0.76}\\
\end{tabular}
\label{tab:correlations}
\end{table*}    

Matching the perception of passengers, the whiteness of effective wheel angles during the drive shows a correlation with DpI, at $r=-0.67$. The whiteness of model outputs $W_\mathsf{{on-policy}}$ and the mean distance from a human trajectory show somewhat weaker correlation with DpI. The measures used here are averages over multiple kilometers and actual danger prediction should happen on a more precise scale. Evaluating whiteness as an online predictor of end-to-end model reliability is outside the scope of this work.

Among the off-policy metrics, $W_{off-policy}$ correlates to a similar degree with intervention frequency as the MAE of steering angles. The difference between the two Pearson correlation coefficients is not significant as per a permutation test. Notice that these two metrics are very different in nature - one measuring the quality and the other temporal stability of predictions. When combining these two metrics via summation after standardization, an even higher correlation with DpI can be obtained, at $r=-0.82$. The improvement over MAE-only correlation is however not statistically significant (permutation test, mean effect size -0.05, $p_{val}$=0.16).

To our knowledge, mean absolute error is a very commonly used off-policy metric for estimating model quality before deployment and for early stopping during model training. MAE has been shown to correlate with driving ability better than multiple other metrics \cite{codevilla2018offline}. Our result suggests that the whiteness of the command sequence generated on an appropriate validation set might serve as a complementary model-selection metric (cf. Discussion).

\section{Discussion}

In the present work we collected a high-quality dataset for the end-to-end road following task in challenging rural roads used for World Rally Championship. This dataset contains driving in narrow and complicated routes during the four seasons of the temperate climate. The measurements of all sensors, including those not used here, across more than 500 km of driving are made publicly available.

On two separate sensory inputs of this dataset, LiDAR-image and frontal camera, we trained models to control the steering of the car. The models were evaluated off-policy and on-policy with speed fixed to 80\% of human-speed. We show that LiDAR-image input as produced by Ouster OS1-128 LiDAR firmware contains sufficient information for road following also in the complex and narrow rally tracks designed to be challenging for humans. The task is not trivial as evidenced by the similarly-trained RGB-image-based models achieving similar performance. Curiously, also models using only the intensity channel of the LiDAR image, information often discarded in point cloud analysis, performed competitively. If this information is sufficient or useful in more complex driving tasks such as highway driving may merit further study. 

The benefits of LiDAR-based driving become apparent when needing to generalize to new conditions. LiDAR images are more similar across weather conditions (cf. Appendix I) and we hypothesize that this allows the entirety of the training data to be useful for driving in all conditions, including those not in the training data. Driving demonstrations from sunny summer days benefit LiDAR-based driving on a dark autumn night, as evidenced by our LiDAR-models being able do drive in the night and to some extent even in the winter. In contrast, RGB-based models can not generalize to night driving. It seems that for a simple RGB-based behavioral cloning approach demonstrations of various traffic situations need to exist in a variety of visually different conditions, increasing the data need. A higher data efficiency of LiDAR-based models would be an interesting property at least for research institutions that cannot boast fleets of cars collecting massive amounts of data daily. 

During night driving LiDAR models can rely on intensity and depth channels which are active sensing and independent of external light sources. Depth channel is also independent of the reflectivity of the surfaces and yields in-distribution values also with snow coverage. While depth alone was proven insufficient for safe end-to-end driving even in training conditions, it may still contribute reliable information to the 3-channel models. Furthermore, we assume that the importance of depth information becomes more apparent in highway and urban driving tasks, where distance with other traffic participants must be maintained.

LiDAR information is often used in its point cloud representation. Here, using image representation allowed us to perform a fair comparison of LiDAR and RGB camera input modalities, as identical methods could be applied. We believe processing LiDAR data in image form can be useful in general, because computer vision is one of the most studied topics in deep learning and many established architectures exist for image processing. Certain architectures are empirically validated to perform various tasks in a reliable manner and methods exist for sensitivity analysis. 

Evaluating autonomous driving systems is complicated because the ability to drive safely can only be measured by deploying the model. When exploring architectures to use, data sampling techniques, or other aspects of the training procedure, one would need to deploy the models to know which techniques work best. This is extremely costly in the real world. If a combination of off-policy metrics could be found that correlates reliably with actual driving ability when deployed, only the more promising models could be selected for testing and evident failures discarded. Here we showed that among sufficiently capable models, the whiteness, i.e. smoothness of generated commands on an appropriate validation set predicted driving ability equally well as the magnitude of errors. We hypothesize that non-smoothness reveals models' uncertainty about the situation -- model reacts differently to very similar inputs. In future work, we propose to evaluate the correlations of other measures of epistemic uncertainty with on-policy performance. Using a more general uncertainty measure carries the benefit of being applicable to a wider range of output modalities, e.g. trajectories and cost-maps. However, these metrics capture only variance and not bias, and a trivial constant model would show perfect stability. Hence, such stability measures should be used in combination with other metrics (e.g. MAE).

\section{Aknowledgements}
This work was supported by the Estonian Research Council grant PRG1604 and collaboration project LLTAT21278 with Bolt Technologies.

\appendices
\onecolumn
\begin{minipage}{\textwidth}
\section{LiDAR and camera image stability across seasons}
\begin{center}
        \includegraphics[width=0.9\columnwidth]{fig/summer_lidar_crop.jpg}
    \includegraphics[width=0.9\columnwidth]{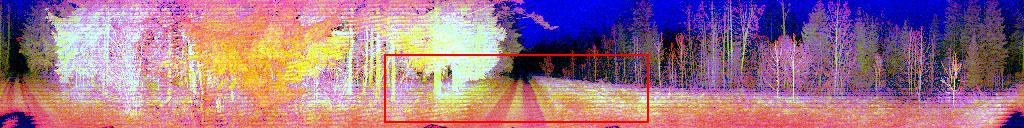}
    \includegraphics[width=0.9\columnwidth]{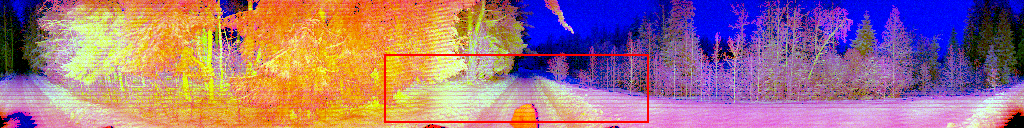}
    \includegraphics[width=0.9\columnwidth]{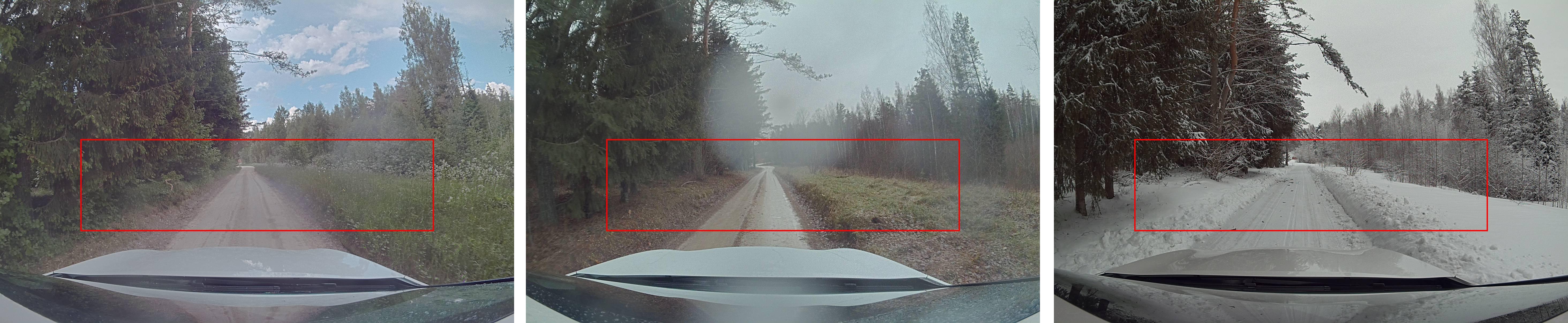}
\end{center}

\scriptsize
Fig. 2. LiDAR and camera images in summer, autumn and winter (from top to down for LiDAR, left to right for camera). The area used for model inputs is marked with red rectangle. In LiDAR images, red channel corresponds to intensity, green to depth and blue to ambient radiation.
\end{minipage}

\clearpage
\begin{minipage}{\textwidth}
\section{Visualization of individual LiDAR channels}
\begin{center}
        (a) Intensity:\\
    \vspace{0.3cm}
    \includegraphics[width=0.9\columnwidth]{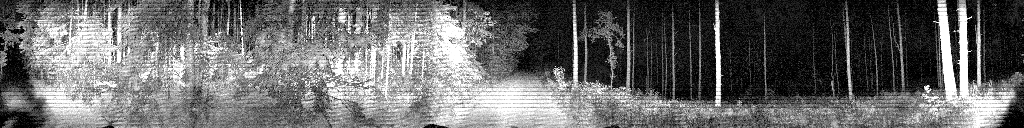}
    \includegraphics[width=0.9\columnwidth]{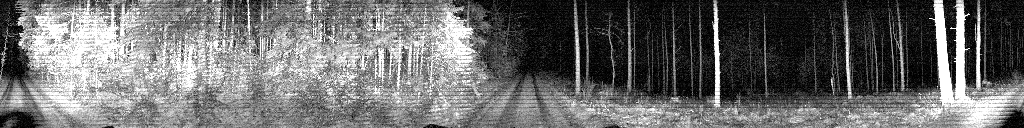}
    \includegraphics[width=0.9\columnwidth]{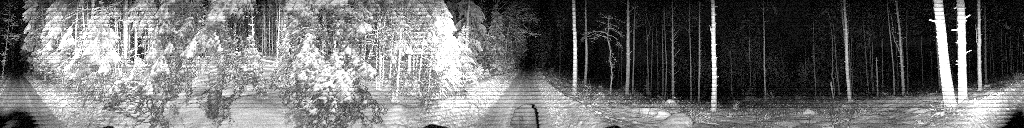}\\
    \vspace{0.2cm}
    (b) Depth:\\
    \vspace{0.2cm}
    \includegraphics[width=0.9\columnwidth]{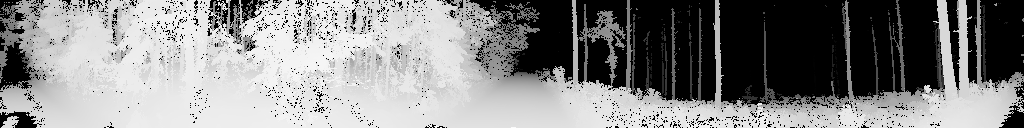}
    \includegraphics[width=0.9\columnwidth]{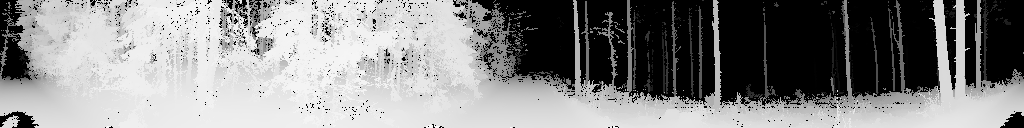}
    \includegraphics[width=0.9\columnwidth]{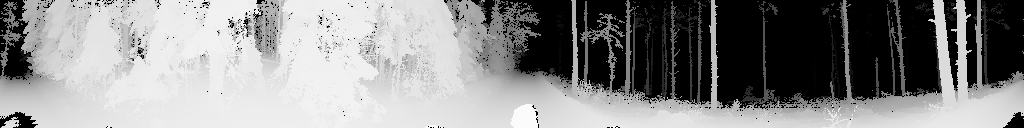}\\
    \vspace{0.2cm}
    (c) Ambient radiation:\\
    \vspace{0.2cm}
    \includegraphics[width=0.9\columnwidth]{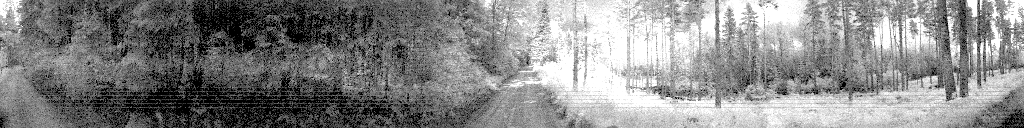}
    \includegraphics[width=0.9\columnwidth]{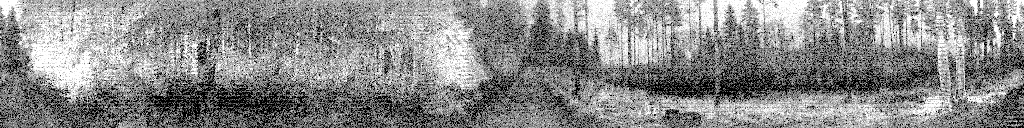}
    \includegraphics[width=0.9\columnwidth]{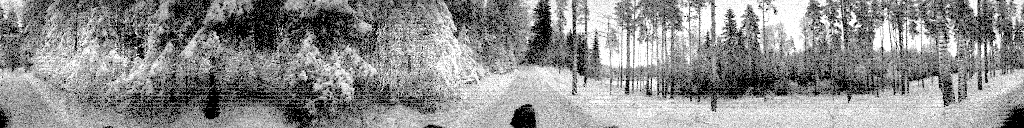}\\
\end{center}
\scriptsize
Fig. 3. LiDAR channels at the same location across the three seasons, in order from top down: summer, autumn, winter. (a) In the intensity channel we see significant difference in how the road itself looks like, while vegetation is surprisingly similar despite deciduous plants having lost their leaves in autumn and winter. (b) Depth image looks stable across seasons, but rather uninformative, as road and low vegetation areas are hard to discern. (c) Ambient radiation images vary strongly in brightness across the seasons, while also displaying strong noise. The noise looks akin to white noise or salt-and-pepper noise and authors do not know its cause.
\end{minipage}

\clearpage
\section{Intervention locations in the winter}
\setcounter{figure}{3}
\begin{figure}[h!]
    \centering
    \includegraphics{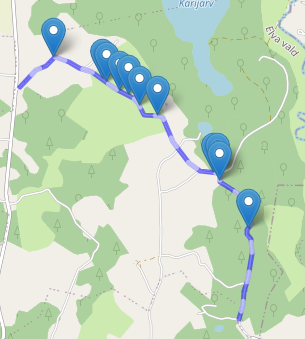}
    \caption{Interventions of a LiDAR v1 model in the winter. The interventions are far more frequent at open fields, whereas the model can handle driving in the forest much better. Also the middle section of the route which contains bushes by the roadside is driven well. }
    \label{fig:my_label}
\end{figure}
\clearpage

\begin{minipage}{\textwidth}
\section{Dataset details}
\scriptsize
\textsc{Table IV. Training dataset. The recording names start with dates in yyyy-mm-dd format, followed by hour. Certain recordings failed to record GNSS locations, but recorded camera and LiDAR feeds and steering, so we can still use them for model training.}
\centering
\vspace{3mm}

\begin{tabular}{c|c|c|c|c|}
 \hline
 & Recording name & Rally Estonia 2021 stage name & Complete & Length (km)\\ 
 \hline
1 & 2021-05-20-12-36-10\_e2e\_sulaoja\_20\_30 & Shakedown & 1 & 6.8 \\
2 & 2021-05-20-12-43-17\_e2e\_sulaoja\_20\_30 & Shakedown, backwards & 0 & GNSS error \\
3 & 2021-05-20-12-51-29\_e2e\_sulaoja\_20\_30 & Shakedown, backwards & 0 & sums to 6.8 km with above  \\
4 & 2021-05-20-13-44-06\_e2e\_sulaoja\_10\_10 & Shakedown & 1 & 6.83 \\
5 & 2021-05-20-13-51-21\_e2e\_sulaoja\_10\_10 & Shakedown, backwards & 0 & 6.14 \\
6 & 2021-05-20-13-59-00\_e2e\_sulaoja\_10\_10 & Shakedown, backwards & 0 & 1.12 \\
7 & 2021-05-28-15-07-56\_e2e\_sulaoja\_20\_30 & Shakedown  & 1 & 6.77 \\
8 & 2021-05-28-15-17-19\_e2e\_sulaoja\_20\_30 & Shakedown, backwards & 0 & 1.43 \\
9 & 2021-06-09-13-14-51\_e2e\_rec\_ss2 & SS12/SS16 & 1 & 22.04 \\
10 & 2021-06-09-13-55-03\_e2e\_rec\_ss2\_backwards & SS12/SS16 backwards & 1 & 24.82 \\
11 & 2021-06-09-14-58-11\_e2e\_rec\_ss3 & SS4/SS8 backwards & 1 & 17.74 \\
12 & 2021-06-09-15-42-05\_e2e\_rec\_ss3\_backwards & SS4/SS8  & 1 & 17.68 \\
13 & 2021-06-09-16-24-59\_e2e\_rec\_ss13 & RE2020 stage, overlaps SS13/17 \& SS3/7 & 1 & 13.96 \\
14 & 2021-06-09-16-50-22\_e2e\_rec\_ss13\_backwards & RE2020 stage, overlaps SS13/17 \& SS3/7 & 1 & 13.99 \\
15 & 2021-06-10-12-59-59\_e2e\_ss4 & RE2020 stage, overlaps SS13/17 \& SS3/7 & 1 & 10.03  \\
16 & 2021-06-10-13-19-22\_e2e\_ss4\_backwards & RE2020 stage, overlaps SS13/17 \& SS3/7 & 1 & 10.14  \\
17 & 2021-06-10-13-51-34\_e2e\_ss12 & RE2020 stage, overlaps with SS2/6 & 1 & 6.9 \\
18 & 2021-06-10-14-02-24\_e2e\_ss12\_backwards & RE2020 stage, overlaps with SS2/6 & 1 & 6.85\\
19 & 2021-06-10-14-44-24\_e2e\_ss3\_backwards & SS4/SS8 & 0 & 16.23 \\
20 & 2021-06-10-15-03-16\_e2e\_ss3\_backwards & SS4/SS8 & 0 & 1.14 \\
21 & 2021-06-14-11-08-19\_e2e\_rec\_ss14 & RE2020 stage, overlaps with SS5/9 & 0 & 6.98 \\
22 & 2021-06-14-11-22-05\_e2e\_rec\_ss14 & RE2020 stage, overlaps with SS5/9 & 0 & 10.77 \\
23 & 2021-06-14-11-43-48\_e2e\_rec\_ss14\_backwards & RE2020 stage, overlaps with SS5/9 & 1 & 18.85 \\
24 & 2021-09-24-11-19-25\_e2e\_rec\_ss10 & SS10/SS14 & 0 & 12.11 \\
25 & 2021-09-24-11-40-24\_e2e\_rec\_ss10\_2 & SS10/SS14 & 0 & 6.06 \\
26 & 2021-09-24-12-02-32\_e2e\_rec\_ss10\_3 & SS10/SS14 & 0 & 3.36 \\
27 & 2021-09-24-12-21-20\_e2e\_rec\_ss10\_backwards & SS10/SS14 backwards & 1 & 23.9 \\
28 & 2021-09-24-13-39-38\_e2e\_rec\_ss11 & SS11/SS15 & 1 & 12.26 \\
29 & 2021-09-30-13-57-00\_e2e\_rec\_ss14 & SS5/SS9 & 0 & 0.93 \\
30 & 2021-09-30-15-03-37\_e2e\_ss14\_from\_half\_way & SS5/SS9 & 0 & 7.89 \\
31 & 2021-09-30-15-20-14\_e2e\_ss14\_backwards & SS5/SS9, backwards & 1 & 19.26 \\
32 & 2021-09-30-15-56-59\_e2e\_ss14\_attempt\_2 & SS5/SS9 & 0 & 19.2 \\
33 & 2021-10-07-11-05-13\_e2e\_rec\_ss3 & SS4/SS8 & 1 & 17.62 \\
34 & 2021-10-07-11-44-52\_e2e\_rec\_ss3\_backwards & SS3/SS7 & 1 & 17.47 \\
35 & 2021-10-07-12-54-17\_e2e\_rec\_ss4 & SS3/SS7 & 1 & 9.16\\
36 & 2021-10-07-13-22-35\_e2e\_rec\_ss4\_backwards & SS3/SS7 backwards & 1 & 9.14\\
37 & 2021-10-11-16-06-44\_e2e\_rec\_ss2 & SS12/SS16 & 0 & GNSS error \\
38 & 2021-10-11-17-10-23\_e2e\_rec\_last\_part & SS12/SS16 paved section & 0 & sums to 12.6km with above \\
39 & 2021-10-11-17-14-40\_e2e\_rec\_backwards & SS12/SS16 backwards & 0 & GNSS error \\
40 & 2021-10-11-17-20-12\_e2e\_rec\_backwards & SS12/SS16 backwards & 0 & sums to 12.6km with above \\ 
41 & 2021-10-20-13-57-51\_e2e\_rec\_neeruti\_ss19\_22 & SS19/22 & 1 & 8\\
42 & 2021-10-20-14-15-07\_e2e\_rec\_neeruti\_ss19\_22\_back & SS19/22 & 1 & 8\\
43 & 2021-10-20-14-55-47\_e2e\_rec\_vastse\_ss13\_17 & SS13/SS17 & 1 & 6.67\\
44 & 2021-10-25-17-06-34\_e2e\_rec\_ss2\_arula\_back & SS2/SS6 & 1 & 12.83\\
45 & 2021-10-25-17-31-48\_e2e\_rec\_ss2\_arula & SS2/SS6 & 1 & 12.82\\
\hline
 \multicolumn{4}{r|}{Total distance:} & 465.89\\
    \end{tabular}

\vspace{1cm}
\textsc{Table V. Validation dataset used for early stopping. The recording names start with dates in yyyy-mm-dd format, followed by hour. The last two recordings are also used in the training set of models overfitted to the evaluation track. Sections of the last two recordings (sections corresponding to the on-policy testing route) were used for obtaining seasonal off-policy metrics for models tested in the autumn.}
\vspace{0.3cm}

\begin{tabular}{c|c|c|c|c|c}
\hline
 & Recording name & Rally Estonia 2021 stage name & Complete & Length (km)& Comment\\ 
 \hline
1 & 2021-05-28-15-19-48\_e2e\_sulaoja\_20\_30 & Shakedown, backwards & 0 & 4.81 & \\ 
2 & 2021-06-07-14-06-31\_e2e\_rec\_ss6 & SS20/SS23 & 0 & 1.45 & \\ 
3 & 2021-06-07-14-09-18\_e2e\_rec\_ss6 & SS20/SS23 & 0 & 2.06 & \\ 
4 & 2021-06-07-14-20-07\_e2e\_rec\_ss6 & SS20/SS23 & 1 & 11.46 & \\ 
5 & 2021-06-07-14-36-16\_e2e\_rec\_ss6 & SS20/SS23, backwards & 1 & 11.41 & \\ 
6 & 2021-09-24-14-03-45\_e2e\_rec\_ss11\_backwards & SS11/SS15 backwards & 1 & 9.83 & used in autumn val. set\\ 
7 & 2021-10-11-14-50-59\_e2e\_rec\_vahi & not RE stage & 1 & 4.97 & used in autumn val. set \\ 
8 & 2021-10-14-13-08-51\_e2e\_rec\_vahi\_backwards" & not RE stage & 1 & 4.9 & used in autumn val. set\\ 
9 & 2021-10-20-15-11-29\_e2e\_rec\_vastse\_ss13\_17\_back & SS13/SS17 & 1 & 7.11 & used in autumn val. set\\ 
10 & 2021-10-26-10-49-06\_e2e\_rec\_ss20\_elva & SS20/SS23 & 1 & 10.91 & used in "overfit", autumn val. set\\ 
11 & 2021-10-26-11-08-59\_e2e\_rec\_ss20\_elva\_back & SS20/SS23 backwards & 1 & 10.89 & used in "overfit", autumn val. set\\ 
\hline
\multicolumn{4}{r|}{Total distance:} & 79.8 &\multicolumn{1}{c}{} \\ 
\end{tabular}

\end{minipage}

\begin{minipage}{\textwidth}
\scriptsize
\centering

\vspace{1cm}
\textsc{Table VI. Winter recordings used for seasonal off-policy metrics computation for models tested in the winter. Only a 4.3km sections from each recording was used, corresponding to the on-policy test route. The recording names start with dates in yyyy-mm-dd format, followed by hour.}
\vspace{0.3cm}

\begin{tabular}{c|c|c|c|c}
\hline
& Recording name & Rally Estonia 2021 stage name & Complete & Length (km)\\ 
\hline
1 & 2022-01-28-14-47-23\_e2e\_rec\_elva\_forward & SS20/SS23 & 1 & 10.52\\
2 & 2022-01-28-15-09-01\_e2e\_rec\_elva\_backward & SS20/SS23 & 1 & 10.82\\
    \end{tabular}
    \label{tab:winter_val}

\vspace{1cm}
\textsc{Table VII. Spring recordings used for seasonal off-policy metrics computation for models tested in the spring. The recording names start with dates in yyyy-mm-dd format, followed by hour.}
\vspace{0.3cm}

\begin{tabular}{c|c|c|c|c}
\hline
& Recording name & Rally Estonia 2021 stage name & Complete & Length (km)\\ 
\hline
1 & 2022-05-04-10-54-24\_e2e\_elva\_seasonal\_val\_set\_forw & SS20/SS23 & 0 & 4.3\\
2 & 2022-05-04-11-01-40\_e2e\_elva\_seasonal\_val\_set\_back & SS20/SS23 & 0 & 4.3\\
\end{tabular}
    \label{tab:spring_val}
    
\end{minipage}

\section{On-policy and off-policy metrics}

\begin{minipage}{\textwidth}
\scriptsize
\centering

\textsc{Table VIII. On-policy test sessions, on-policy metrics recorded and the corresponding off-policy metrics from the same track in same season. These values serve as the basis for correaltion calculations between distance per intervention (DpI) and other metrics. DpI for tests with 0 interventions was set to 10 km. Combined refers to the sum of $MAE_{steer}$ and $W_{off-policy}$, both standardized to zero mean and standard deviation one.}
\vspace{0.3cm}

\begin{tabular}{p{3cm}|c|c|c|c|c||c|c|c}
Model (session) & DpI & $MAE_\mathsf{{trajectory}}$ & Failure rate & $W_{effective}$& $W_\mathsf{{on-policy}}$ & $MAE_{steer}$ & $W_{off-policy}$ & Combined\\

\hline
Camera v1 BGR (Nov) &	665.00m & 0.2715m & 2.82\% & 109.99°/s &41.36°/s &7.23° &99.00°/s &	3.703541\\
Camera v2 BGR (Nov) &743.87m &0.2906m &3.31\% &58.91°/s &27.27°/s &6.91° &74.50°/s 	&2.577991\\
Camera v3 BGR (Nov) &685.03m &0.2526m &1.54\% &94.10°/s &35.42°/s &7.46° &90.50°/s &3.633461\\
Camera overfit BGR (Nov) &	1185.92m &	0.2707m 	&4.83\% &	68.16°/s &	28.36°/s &	6.86°& 	116.00°/s &	4.038257\\
LiDAR v1 (Nov)&	4221.26m &	0.2164m &	0.42\% &	60.72°/s &	22.96°/s &	5.93°& 	52.40°/s &	1.202971\\
LiDAR v2 (Nov) &4232.96m &	0.2395m& 	0.98\% &	40.79°/s &	17.71°/s &	6.05° &	30.10°/s &	0.509354\\
LiDAR v3 (Nov) 	&2810.78m &	0.2536m & 2.18\%& 	56.74°/s &	18.84°/s &	5.88°& 	41.00°/s& 0.745176\\
LiDAR overfit (Nov)& 	10000.00m &	0.2599m &	4.38\% &	38.54°/s& 	19.21°/s& 	4.02°& 27.30°/s& -1.221416\\
Camera v1 RGB (May) &4232.17m &	0.2309m &0.96\% &65.03°/s &	24.63°/s &	6.26° &	63.20°/s& 	1.745492\\
Camera v2 RGB (May) &2090.79m &	0.2382m &0.69\% &	107.73°/s& 	33.46°/s &	6.72°& 	54.70°/s 	&1.910144\\
Camera v3 RGB (May) &1198.55m &	0.2403m	&0.66\% &	67.55°/s &	29.70°/s &	6.68° 	&63.66°/s& 	2.076987\\
Camera overfit RGB (May)&	2829.71m &	0.2453m& 	1.39\% &	65.94°/s& 	23.07°/s &	5.96°& 	59.91°/s &	1.405479\\
LiDAR v1 shifted (May) &	436.15m &	0.2649m &1.52\% &	49.76°/s& 	23.66°/s &	7.54°& 	64.41°/s &	2.775023\\
LiDAR v2 shifted (May) &	481.37m &	0.2673m &3.52\% &52.64°/s &	27.25°/s &8.03° &59.80°/s& 	3.155804\\
LiDAR v3 shifted (May) &	251.56m &	0.2786m &	3.20\% &92.61°/s &40.73°/s &9.64° &90.70°/s& 	5.404183\\
LiDAR overfit shifted (May) &	603.82m &	0.2783m &4.69\% &56.62°/s &31.01°/s &7.88°	&73.70°/s &3.388144\\
LiDAR v2 day 2 (Nov) &	10000.00m &	0.2193m &1.97\% &33.89°/s &	19.17°/s &6.05°& 	30.10°/s &	0.509354
\end{tabular}

\end{minipage}

\section{Qualitative observations of sensitivity}
As a critique to behavioral cloning, we experienced that the models were highly sensitive to shifts in inputs. First of all, we accidentally performed experiment of feeding BGR images to models trained with RGB images. These models were able to drive with BGR input, but clearly worse than with clean RGB input. For a human, the scene would still be easily understandable after switching blue and red colors, especially with dim light and cloudy skies as during autumn testing. The comparison of RGB-models' performance using BGR and RGB inputs can be seen in Table VIII, albeit not in similar conditions (Nov and May sessions, respectively).

Similarly, we performed unplanned test with LiDAR sensor moved from its original location in the center of the roof to the front of the roof. The performance of LiDAR model suffered considerably despite our attempts to fix it by changing the cropped area from the surround-view LiDAR-image. The comparison of LiDAR performance with original and shifted location can be seen in Table VIII, Nov and May sessions respectively. 

More worrying is the fact that changing the location of crop by only one pixel to the left or right resulted in perceivable biases in on-policy driving behavior. Also in vertical direction, few pixels difference in crop height clearly mattered. Indeed, logically, a crop more from the left should result in turning more towards the right and in a sequential decision-making task this effect can accumulate over time. However, the fact that just one pixel difference can bias the model to position itself differently on the road and struggle with certain turns, is worrying. We conclude that at least the end-to-end approach used here is extremely sensitive to changes in sensor data.



\twocolumn
\bibliographystyle{IEEEtran}
\bibliography{IEEEabrv,references}

\end{document}